\documentclass[conference]{IEEEtran}
\IEEEoverridecommandlockouts
\usepackage{cite}
\usepackage{amsmath,amssymb,amsfonts}
\usepackage{algorithmic}
\usepackage{graphicx}
\usepackage{textcomp}
\usepackage{xcolor}
\def\BibTeX{{\rm B\kern-.05em{\sc i\kern-.025em b}\kern-.08em
    T\kern-.1667em\lower.7ex\hbox{E}\kern-.125emX}}
\begin{document}
\title{Gait analysis with curvature maps:
\\A simulation study\\}
\author{\IEEEauthorblockN{Khac Chinh Tran}
\IEEEauthorblockA{\textit{Information Technology Faculty} \\
\textit{Danang University of Technology and Science}\\
Danang, Vietnam \\
chinhpfiev@gmail.com}
\and
\IEEEauthorblockN{Marc Daniel}
\IEEEauthorblockA{\textit{Laboratoire d'Informatique et des Systèmes} \\
\textit{Aix-Marseille Université}\\
Marseille, France \\
marc.daniel@univ-amu.fr}
\and
\IEEEauthorblockN{Jean Meunier}
\IEEEauthorblockA{\textit{Department of Computer Science} \\
\textit{University of Montréal}\\
Montréal, Canada \\
meunier@iro.umontreal.ca}
}
\maketitle
\begin{abstract}
Gait analysis is an important aspect of clinical investigation for detecting neurological and musculoskeletal disorders and assessing the global health of a patient. In this paper we propose to focus our attention on extracting relevant curvature information from the body surface provided by a depth camera. We assumed that the 3D mesh was made available in a previous step and demonstrated how curvature maps could be useful to assess asymmetric anomalies with two simple simulated abnormal gaits compared with a normal one. This research set the grounds for the future development of a curvature-based gait analysis system for healthcare professionals.
\end{abstract}

\section{Introduction}
Assessment of mobility, gait and balance is an important aspect of clinical practice to detect and assess neurological and musculoskeletal disorders of a patient [1]. Current methods use questionnaires, functional tests or more complex equipment requiring markers and trained specialists. Video analysis with a standard (RGB) or a depth (RGBD) camera constitutes an interesting alternative where an appropriate model of the subject (e.g., skeleton) is fitted to the subject’s body for gait analysis. These systems are acceptable for gait analysis with the advantage of being much more affordable and simpler to set up [2]. Notice that most researchers working with these systems limit their analysis to measurement of the joint angles and positions of the fitted skeleton. Here we focus on directly extracting relevant (and complementary) information from the reconstructed 3D body surface (no skeleton) under the form of curvature maps. For instance, real-time 3D body reconstruction can be achieved using the point clouds provided by a depth camera in front of a subject walking on a treadmill (e.g. [3]). To extract curvature maps, a 3D surface mesh is fitted to the point cloud using an appropriate algorithm [4, 8]. In this paper we assume that the 3D mesh was made available hand as input; our goal is thus to demonstrate how curvature maps can be extracted from the body 3D mesh and how they could be useful for gait analysis. For that purpose, we conducted a study on three simulated gaits, including one normal gait and two abnormal gaits. This work could contribute to set the grounds for the future development of a curvature-based gait analysis system for healthcare professionals.

\section{Basic concepts of curvature}
There are many types of curvatures but four of them are of particular interest for researchers: Gaussian curvature, mean curvature, absolute curvature and root mean square curvature.
\subsection{Principal curvatures and the four basic curvatures}
All four curvatures can be calculated at a given point on a surface based on the principal curvatures ($\kappa_1$ and $\kappa_2$) which are defined as the maximum and minimum values of the curvature of the curve resulting from the intersections of a normal plane with the surface at that point.
\paragraph{Gaussian curvatures ($\textbf{\textit{K}}$)}
In differential geometry, $\textbf{\textit{K}}$ of a surface is the product of the principal curvatures, $\kappa_1$ and $\kappa_2$, at the given point:
\begin{equation}
    K=\kappa_1\kappa_2
\end{equation}

As a result, the sign of the Gaussian curvature can be used to characterise the surface.
For example, when the Gaussian curvature is positive the surface is dome shaped, when negative the surface is saddle shaped.
\paragraph{Mean curvatures ($\textbf{\textit{H}}$)}
The mean curvature $\textbf{\textit{H}}$ at a point is defined as the average of the principal curvatures:
\begin{equation}
    H=\frac{1}{2}(\kappa_1 + \kappa_2)
\end{equation}

\paragraph{Absolute curvatures ($\textit{\textbf{K\textsubscript{abs}}}$)}
The total absolute curvature is defined by the sum of the absolute values of the principal curvatures:
\begin{equation}
    K_{abs}=|\kappa_1| + |\kappa_2|
\end{equation}

\paragraph{Root mean square curvatures ($\textit{\textbf{K\textsubscript{rms}}}$)}
Similarly to the absolute curvature, the $\textit{\textbf{K\textsubscript{rms}}}$ is a characteristic value for evaluating surface flatness. It is determined by the formula:
\begin{equation}
    K_{rms} = \frac{\sqrt{\kappa_1^2+\kappa_2^2}}{2}
\end{equation}
\subsection{Computing discrete curvatures}
In practice, we need to compute discrete curvatures from a 3D mesh (e.g. reconstructed from a 3D point cloud generated with a depth camera). An effective way to do it is through the discrete Laplace-Beltrami operator [7]. It calculates curvature from the mesh surface of the object instead of principal curvature. With a 3D mesh surface $\textbf{\textit{S}}$ whose vertices are $v_i$, through the discrete Laplace-Beltrami operator [7], we can simply compute the mean curvature as:
\begin{equation}
    H = \frac{1}{2}||\Delta_S v_i||
\end{equation}

On the other hand, if we suppose that $\theta_j$ is the angle created by 2 vectors $\Vec{v_i v_j}$ which $v_j$ is a neighbor of $v_i$ and  $A_{v_i}$ can be simply a $\frac{1}{3}$ of the areas of the triangles that form this ring, the Discrete Gaussian Curvature is defined as:
\begin{equation}
    K = (2\pi - \sum_j \theta_j)/A(v_i)
\end{equation}

Knowing the mean curvature and Gaussian curvature, we can compute  $\kappa_1$, $\kappa_2$ and $K_{rms}$ base on these curvatures. 

Notice that the authors know that formulas (5) and (6) have some limitations but a complete analysis of discrete curvatures is beyond the scope of this paper.
\section{Method}
In this section we present the method used for the 3D simulation of a realistic 3D mesh of a walking human with normal and abnormal gaits.
\subsection{3D model of the human body}
To conduct our study we used MakeHuman [9] to simulate a 3D model of a 25-year-old male subject, 173 cm tall.
\subsection{Animation of the human body}
All walks, whether normal or abnormal, are a repetitive process of gait cycles. A gait cycle can be divided into two halves corresponding to the left and right half of the body [5]. Furthermore, each half cycle can also be divided into four basic postures including: contact, low, passing and high (see Fig. 1). Each gait has its own properties but all gaits must include these four types of posture. The animation of the simulated human body was done with Blender [10]

\begin{figure}[htbp]
\centerline{\includegraphics[scale=0.4]{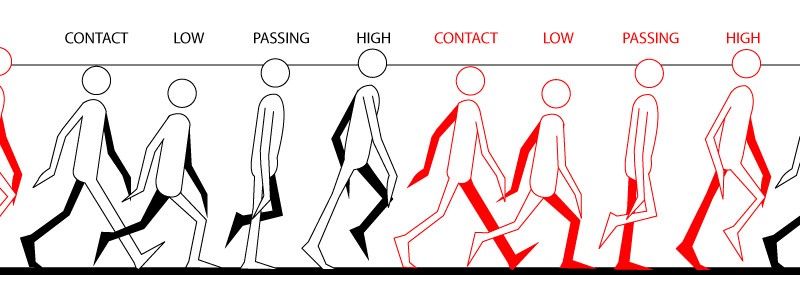}}
\caption{The walking cycle [6].}
\label{fig}
\end{figure}

Each posture of walking can typically be identified according to the joints of the human body. For instance, we can notice that the knee and elbow joints are constantly active during walking to help the body move forward and balance. Therefore, we can expect that the curvature maps in these areas will change dramatically and could characterize the person's gait through the 8 postures in the walking cycle.

\begin{figure}[htbp]
\centerline{\includegraphics[scale=0.6]{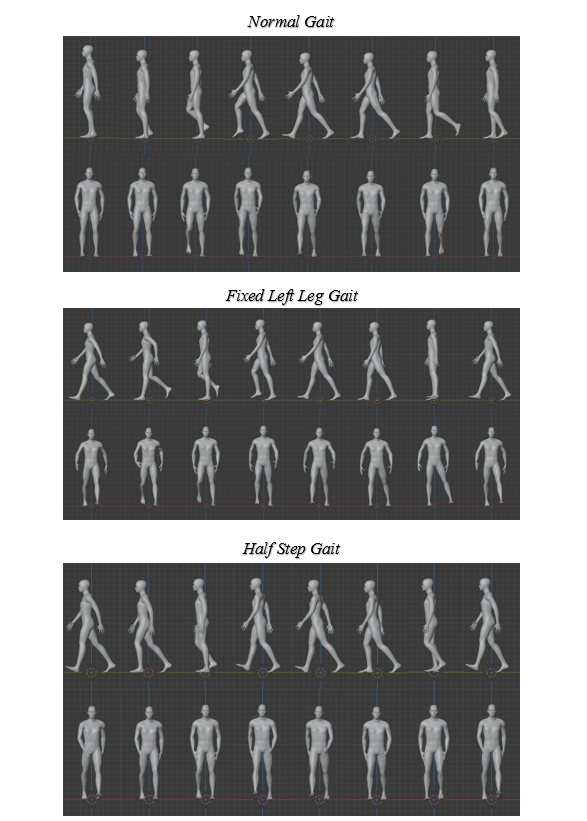}}
\caption{Side view and front view of simulation gaits.}
\label{fig}
\end{figure}

In addition to assessing the change of curvature in a region of interest during the walk, we can compare the results between normal and abnormal gaits to demonstrate significant differences. We can evaluate an interesting curvature area based on three criteria:
\begin{itemize}
    \item The magnitude of the curvature in the area.
    \item The size of the area with a curvature of interest.
    \item The location of the area with an interesting curvature.
\end{itemize}
To evaluate all these factors, we conducted a simulation study on three gaits, including a normal gait and two abnormal gaits which are described as follows (see Fig. 2):
\paragraph{Normal}
Using Blender [10] we simulated a normal walk with a step length of 65 cm. This simulated walk is exactly symmetrical (left-right). The actual gait measured in healthy individuals is also quite (but not perfectly) symmetrical. However, this symmetry is often significantly compromised in the pathological gait.
\paragraph{Locked left knee}
When we lock one leg (the left one in this case), the knee area of this leg is always straightened. This means that the curvature does not change during walking. As a result, for moving forward, the subject is forced to extend the stride on the side. In other words, the left leg trajectory forms a curve instead of straight line (see Fig. 2).
\paragraph{Half step}
Here, the right foot is always in front and the left foot is always behind using half steps for each side. In this case, the right leg has a slightly greater change in curvature than the left leg. Furthermore the front knee has to be lifted higher than the behind knee on average. Therefore, the left and right knee areas show differences in their coordinates for corresponding postures (see Fig. 2). 

Notice that in these simulations, the abnormal gaits only affected the lower limbs of the body.

\subsection{Curvature display on the body}
Discrete curvature maps are computed (section II.B) and displayed as a dynamic color map that changes from blue (negative) to green (zero) and to red (positive) depending on the magnitude of the specific curvature at a certain point on the human body during the motion (see Fig. 4 and 5).

\section{Results and Discussion}
\subsection{Fluctuations of curvatures}
To illustrate the quantitative change of curvature, we focused on one interesting area, which is the knee. We collected data from the knee section by looking at its anthropological position in the area of about 25-30\% of body height. Points with maximum Gaussian curvature were considered to calculate the center point of the knee area. Then, all the points within a 5 cm radius were used to calculate the average curvature within the knee area as a function of time for the four basic curvatures described above. 

The results are presented in Fig. 3 and show the magnitudes of the four curvatures in the left and right knee areas for two walking cycles. Because the subject was a humanoid simulation, the set of points in the knee area was not affected by noise for the two cycles. From theses results, we can draw a few comments as follows:

\begin{itemize}
\item Because of its specific calculation emphasizing dome-like features, Gaussian curvature (blue curve) has high cyclic oscillation in the knee area compared to the other 3 types of curvature. We observed that this was the best type of curvature for showing interesting characteristics of the body during gait analysis. For this reason, We will use this curvature to illustrate curvature maps for the rest of the paper.
\item For a normal gait (first row in Fig. 3), the curvature graphs of the two knees are exactly the same, they only deviate from each other by half a cycle as expected.
\item The right knee area (second column in Fig. 3) for all three gait types displays similar behaviour. The extreme points appear usually in the same key frame. This is understandable because the subject's right foot is relatively free to move in all three gait types. The difference only occurs by the nature of the constraints on the body motion when walking abnormally and walking normally.
\item The left knee curvature maps for the locked left leg (first column, second row) is much more stable. This was expected since in theory, the curvature values of the left knee should be almost constant. However, the act of extending the legs toward the side affected somewhat the detected area due to imperfect detection causing small curvature fluctuations.
\end{itemize}

\begin{figure}[htbp]
\centerline{\includegraphics[scale=0.75]{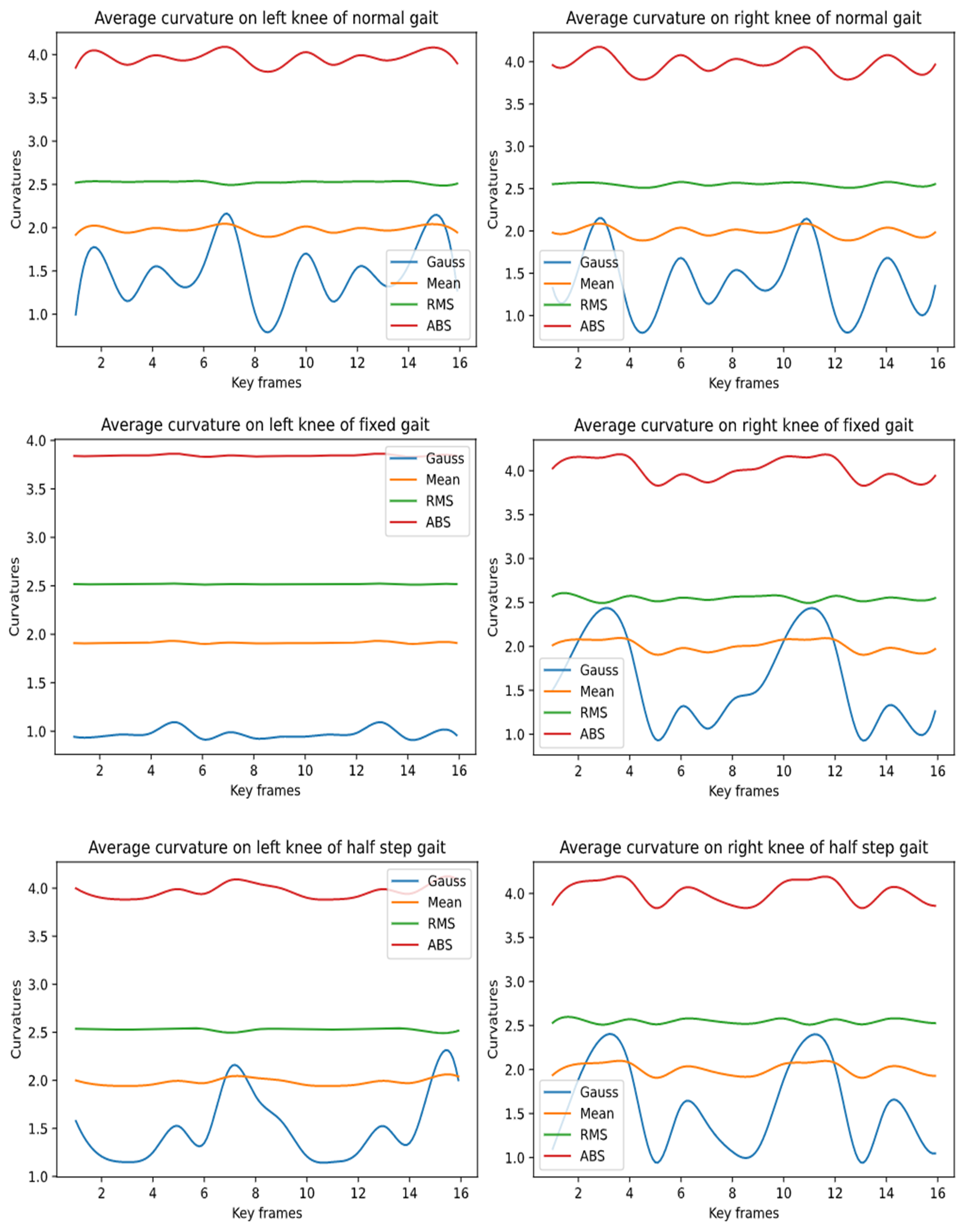}}
\caption{Statistics of average curvature of the knee areas. The 16 key frames of the horizontal axis correspond to the 8 postures of Fig. 1 repeated two times (2 gait cycles).}
\label{fig}
\end{figure}

In summary, we can infer that if the curvature graphs of both knees are similar and only deviate from each other by half a cycle, then the person is walking normally with a symmetric gait. Otherwise, further investigations are needed to establish the nature of the gait disorder.

\subsection{Left-Right Symmetry}

Assuming that a normal walk is nearly symmetrical because of the equal role of each leg for walking (except for their half-a-cycle offset), we propose to evaluate the whole curvature map of the body by showing dynamically the left and right halves together for the same corresponding posture by aligning the postures that were originally offset by half a cycle. This means that the right contact posture is displayed with the left contact posture for comparison purpose and so on. An example is shown in Fig. 4 for the passing posture with the Gaussian curvature.
\begin{figure}[htbp]
\centerline{\includegraphics[scale=0.4]{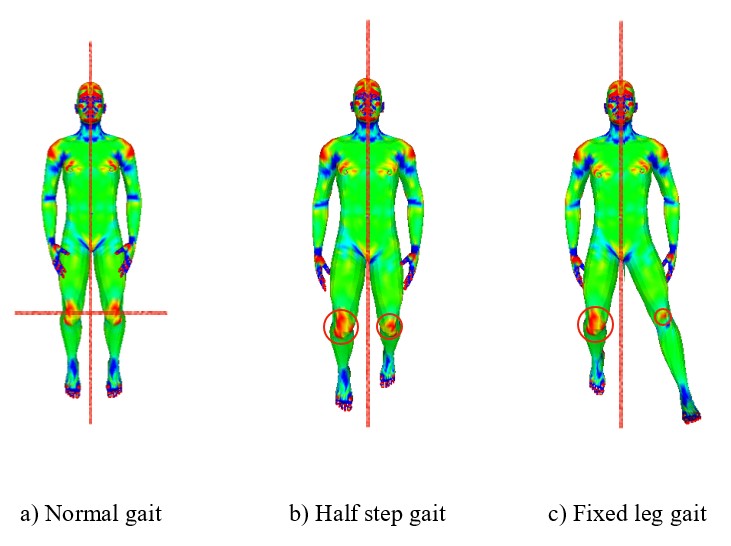}}
\caption{Left-Right Symmetry (Passing pose - Gaussian curvature).}
\label{fig}
\end{figure}
We can see that the curvature map is symmetrical with respect to the vertical red axis for the two halves of the body for a normal walk. As expected, for the two types of abnormal gait, at the passing pose, there are noticeable differences at the level of the legs between the two halves of the body. The curvature in the area of the right knee of both abnormal gaits is similar to the normal gait, but the amplitude of curvature for the left knee area is less, in other words the left knee is less bent than the right one. Moreover, the position of the knees is uneven in this passing pose.

In a nutshell, for these simulations, if we measure the position and curvature of the knee area in two postures, half-a-cycle apart, we could predict whether that gait is in the normal or abnormal class. Of course, other areas could be of interest for other types of abnormal gaits. Note that in the presence of noise, the average of several walking cycles can provide good curvature maps.

\subsection{Average curvature maps}
Another method to investigate the symmetry of curvature maps is to average the curvature across all frames of a gait cycle instead of studying one posture at a time. This way, a single image reveals the symmetry of the whole gait cycle.

\begin{figure}[htbp]
\centerline{\includegraphics[scale=0.5]{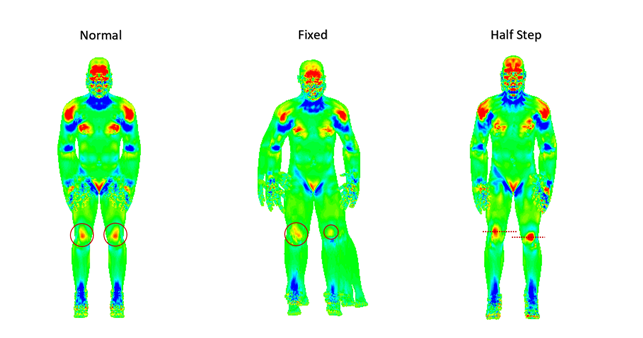}}
\caption{Average curvature maps.}
\label{fig}
\end{figure}

Although the 3D model does not show many noisy points, in practice when a 3D mesh is computed form a point cloud obtained with a depth camera the noise could be an issue, the average of curvature maps could be extremely effective in this case which is another advantage of this mapping method.

From the color map shown in Fig. 5, we can see that some less interesting details are reduced by calculating the average curvature. Interesting positions such as knees, shoulders, wrists and ankles are highlighted.
The calculation of the average curvature map not only significantly reduce the noise that exists on the object surface, it can also show the characteristic properties of the gait. Similarly to the previous analysis, the normal walk shows the expected symmetry of the curvature map with respect to the vertical axis. 

The locked leg simulation exhibited low value of curvature at the left knee level due to the locked knee but also because of the averaging of several postures where the left leg deviates to the left side. For the other abnormal walk, we observed an height difference between the knees. This happens because the front leg always has to lift higher than the behind leg. 

As a result, each gait gives an average curvature map of a completely different nature. This demonstrates the potential of this method to assess gait anomalies and could assist the clinicians in recognizing a particular disorder during walking.

\section{Conclusion}

In this work, our goal was to demonstrate the potential of curvature maps for gait analysis. These maps could be computed on 3D surface meshes of the body adjusted to point clouds provided by a depth camera in front of a subject walking on a treadmill. We focused specifically on how curvature maps can be extracted from the body 3D mesh and how they could be useful to detect gait anomalies. For that purpose, we simulated a normal gait and two abnormal gaits and proposed different methods to quantitatively assess and display the curvature maps in order to reveal gait anomalies. This approach is new, and the results are preliminary. In the future, comparison with other conventional methods with various and more realistic gaits will be needed to better assess its potential benefits. We believe that this work could foster further development of a curvature-based gait analysis system for healthcare professionals.

\end{document}